%% file: main.tex
\documentclass[final]{cvpr}

\makeatletter
\@namedef{ver@everyshi.sty}{}
\makeatother
\usepackage{tikz}

\usepackage{svg}
\usepackage{times}
\usepackage{epsfig}
\usepackage{graphicx}
\usepackage{amsmath}
\usepackage{amssymb}

\usepackage[pagebackref=true,breaklinks=true,colorlinks,bookmarks=false]{hyperref}

\def\cvprPaperID{1138} 

\begin{document}

\title{Adversarial collision attacks on image hashing functions}

\author{Brian Dolhansky\\
Facebook AI\\
{\tt\small bdol@fb.com}
\and
Cristian Canton Ferrer\\
Facebook AI\\
{\tt\small ccanton@fb.com}
}

\maketitle

\begin{abstract}
    Hashing images with a perceptual algorithm is a common approach to solving duplicate image detection problems. However, perceptual image hashing algorithms are differentiable, and are thus vulnerable to gradient-based adversarial attacks. We demonstrate that not only is it possible to modify an image to produce an unrelated hash, but an exact image hash collision between a source and target image can be produced via minuscule adversarial perturbations. In a white box setting, these collisions can be replicated across nearly every image pair and hash type (including both deep and non-learned hashes). Furthermore, by attacking points other than the output of a hashing function, an attacker can avoid having to know the details of a particular algorithm, resulting in collisions that transfer across different hash sizes or  model architectures. Using these techniques, an adversary can poison the image lookup table of a duplicate image detection service, resulting in undefined or unwanted behavior. Finally, we offer several potential mitigations to gradient-based image hash attacks.
\end{abstract}
\input{intro}

\input{background}

\input{method}

\input{experiments}

\input{mitigations}

{\small
\bibliographystyle{ieee}
\bibliography{main}
}

\end{document}


\title{
Supplementary Material:
\\Hash Attacks
}

\author{First Author\\
Institution1\\
Institution1 address\\
{\tt\small firstauthor@i1.org}
\and
Second Author\\
Institution2\\
First line of institution2 address\\
{\tt\small secondauthor@i2.org}
}

\maketitle
\section{Baseline hash results}
\begin{table*}[ht!]
\centering
\footnotesize
\begin{tabular}{|l|l|rrrr|}
\hline
 Model           & Hash           &   Top-1-acc &   Top-5-acc &   Top-10-acc & Coll. rate   \\
\hline
 -               & ahash\_64       &       0.584 &       0.603 &        0.611 &        0.039 \\
 -               & ahash\_144      &       0.607 &       0.619 &        0.625 &        0.009 \\
 -               & ahash\_256      &       0.611 &       0.622 &        0.628 &        0.002 \\
 -               & dhash\_64       &       0.612 &       0.625 &        0.63  &        0.001 \\
 -               & dhash\_144      &       0.609 &       0.616 &        0.62  &        0     \\
 -               & dhash\_256      &       0.617 &       0.628 &        0.632 &        0     \\
 -               & phash\_64       &       0.608 &       0.62  &        0.625 &        0     \\
 -               & phash\_144      &       0.616 &       0.625 &        0.629 &        0     \\
 -               & phash\_256      &       0.614 &       0.621 &        0.624 &        0     \\
 alexnet         & pca\_median\_64  &       0.567 &       0.662 &        0.695 &        0     \\
 alexnet         & pca\_median\_128 &       0.701 &       0.771 &        0.795 &        0     \\
 alexnet         & pca\_median\_256 &       0.77  &       0.827 &        0.847 &        0     \\
 resnet18        & pca\_median\_64  &       0.57  &       0.675 &        0.713 &        0     \\
 resnet18        & pca\_median\_128 &       0.774 &       0.843 &        0.865 &        0     \\
 resnet18        & pca\_median\_256 &       0.868 &       0.913 &        0.926 &        0     \\
 resnet50        & pca\_median\_64  &       0.567 &       0.682 &        0.723 &        0     \\
 resnet50        & pca\_median\_128 &       0.766 &       0.845 &        0.871 &        0     \\
 resnet50        & pca\_median\_256 &       0.864 &       0.917 &        0.933 &        0     \\
 resnet101       & pca\_median\_64  &       0.581 &       0.7   &        0.742 &        0     \\
 resnet101       & pca\_median\_128 &       0.779 &       0.857 &        0.882 &        0     \\
 resnet101       & pca\_median\_256 &       0.869 &       0.92  &        0.936 &        0     \\
 efficientnet-b3 & pca\_median\_64  &       0.617 &       0.72  &        0.757 &        0.001 \\
 efficientnet-b3 & pca\_median\_128 &       0.788 &       0.858 &        0.881 &        0     \\
 efficientnet-b3 & pca\_median\_256 &       0.876 &       0.921 &        0.934 &        0     \\
 efficientnet-b5 & pca\_median\_64  &       0.521 &       0.63  &        0.671 &        0.006 \\
 efficientnet-b5 & pca\_median\_128 &       0.721 &       0.801 &        0.829 &        0     \\
 efficientnet-b5 & pca\_median\_256 &       0.838 &       0.891 &        0.908 &        0     \\
 efficientnet-b7 & pca\_median\_64  &       0.529 &       0.64  &        0.682 &        0.008 \\
 efficientnet-b7 & pca\_median\_128 &       0.73  &       0.809 &        0.836 &        0     \\
 efficientnet-b7 & pca\_median\_256 &       0.847 &       0.897 &        0.913 &        0     \\
\hline
\end{tabular}
\caption{Baseline nearest-neighbor classification accuracy for all 256-bit hashes, as well as incidental collision rates.}
\label{tab:baselines}
\end{table*}


%% file: intro.tex
\section{Introduction}


\begin{figure}[ht]
    \centering
    \includegraphics[width=\columnwidth]{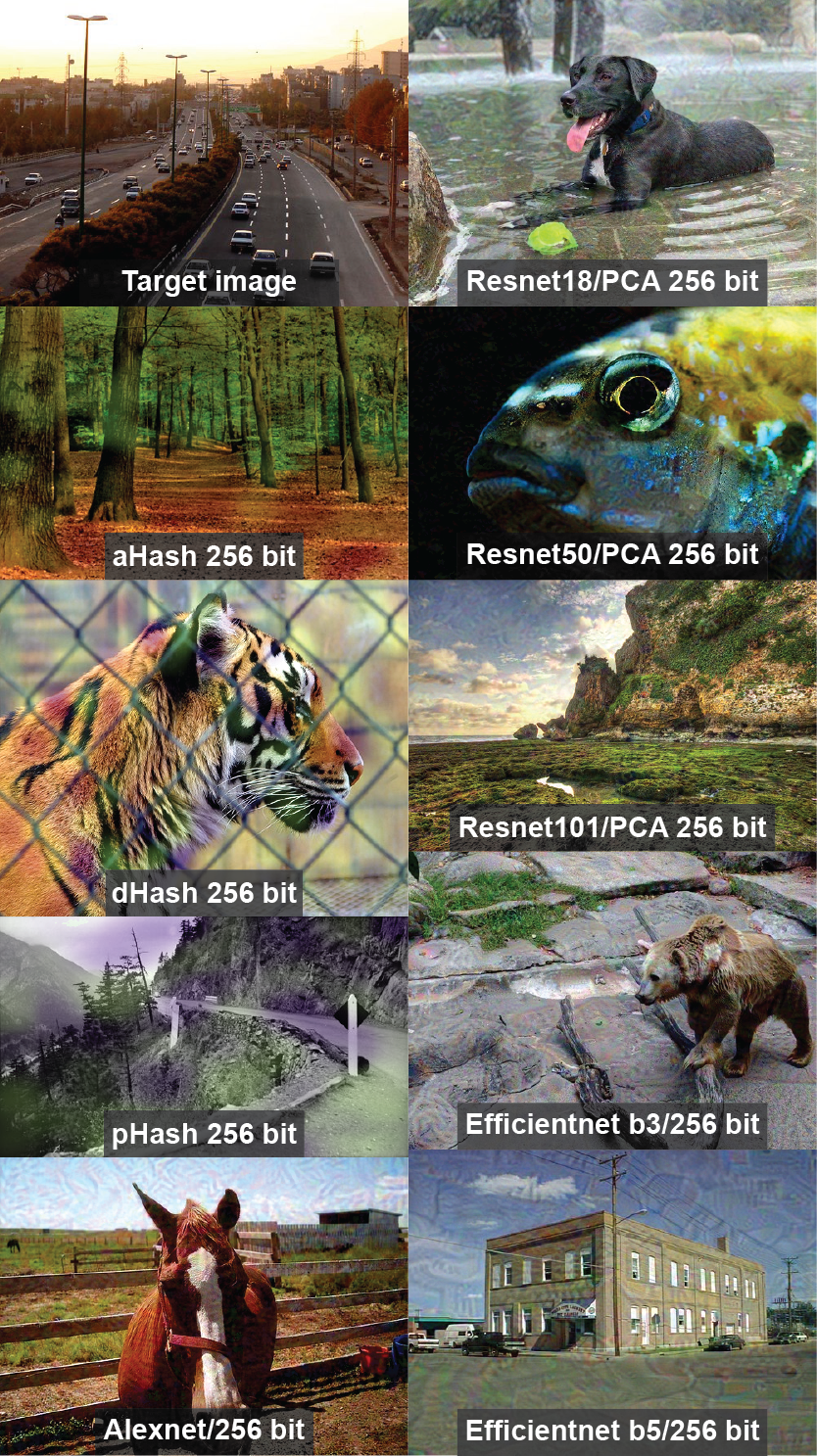}
    \caption{All images in this figure have the same image hash as the top-left image.}
    \label{fig:collage}
    \vspace{-10pt}
\end{figure}

Adversarial attacks on machine learning algorithms have received widespread recent attention, as some of the studied attacks could potentially have dire real-world consequences. In particular, computer vision models have been shown to be susceptible to imperceptible changes to an image. These changes are generally known as adversarial perturbations, and depending on whether or not an attacker has full or limited access to a model, they seem to "fool" nearly all models that make predictions over raw pixels~\cite{moosavi2017universal,Szegedy2013}.

Image hashing algorithms are used widely in critical production systems, and are used to detect anything from copyright violations to misinformation~\cite{Integrity2020}. These algorithms allow fast lookup and retrieval of duplicate or near-duplicate images, and thus provide a cheap way to find similar images among a haystack of billions~\cite{babenko2016efficient}. While many hashes are computed with deep networks (and thus are potentially susceptible to adversarial perturbations), for better scaling, other systems utilize cheaper non-learned algorithms~\cite{photodna}. The choice of which algorithm to use depends on the application, but in general, once an algorithm is chosen it remains fixed to avoid unnecessary re-computation of the image hash database, or because access to the original images is no longer possible.

However, image hashes, like other hashes, are vulnerable to \textit{collisions}, where two different images map to the same hash. In fact, because of some inherent properties of image hashing functions, these algorithms are particularly susceptible to a \textit{collision attack} (formally known in cryptography as a \textit{preimage attack}), where an adversary deliberately modifies an input image so that it maps to the same hash as some other unrelated image. The applications of this type of attack can include \textit{evasion}, where an adversary changes the hash of an undesirable image to exactly match that of a benign image, or a \textit{poisoning attack via collision}, where the hash is changed but with the intention of polluting some hash lookup table, thus causing a duplicate image detection system to behave in an unknown or even harmful manner. 

While previous research has examined the efficacy of targeted attacks designed to collide classes in a single dataset~\cite{li2020towards}, we instead focus on the much stronger criteria of creating exact image hash collisions which are dataset-agnostic. Furthermore, we examine various attack vectors from a holistic standpoint. An end-to-end image hashing algorithm contains multiple steps, each of which are specific to the the task of generating an image hash, but each of which provide an exploitable weak point. In addition, creating a collision at an early stage in the hashing pipeline will cause that collision to cascade to the final hash, potentially increasing the effectiveness of gray- or black-box transfer attacks. As many algorithms consist of similar early steps (such as resizing the image to a fixed size, or utilizing similar low-level learned filters in deep networks), attacking an earlier step allows an adversary to avoid having to know any details about the end stages of the hashing pipeline. For example, if a collision is created within a deep network that produces embeddings used to create a hash, the specific hashing algorithm parameters become irrelevant.

We examine the susceptibility of image hashing algorithms to gradient-based adversarial attacks, but with several important novel contributions. (1) We consider the entire hashing system, including the image resizing step, which allows perturbations of full-resolution and non-cropped images; (2) we examine \textit{interior surjective} attacks in the regime of deep networks, where we seek to create collisions not only at the output of the network, but at the intermediate output level; (3) we demonstrate that gradient-based adversarial attacks are not only effective against deep hashes, but also effective against the type of "shallow", non-learned hashes that are often used in large-scale detection systems; (4) we show that all of these attacks are effective not only in a white-box setting, but can transfer between a variety of models and methods; (5) we use these results to determine the effectiveness of several mitigation strategies.

%% file: background.tex
\section{Background}
\par Duplicate image detection is a well-studied task~\cite{chum2008near,dong2012high,ke2004efficient}, with many varying solutions and applications. One may wish to find image duplicates to protect copyright~\cite{venkatesan2003system}, to remove unwanted or illegal images from a platform~\cite{photodna,Integrity2020}, or for forensic analysis~\cite{lu2010forensic,steinebach2012forbild}. One practical solution to this problem is to check a probe image against a bank of known images and subsequently performing an action such as removal or flagging if a match is found. There are a variety of methods that can be used to perform duplicate detection, but for many large-scale systems, binary perceptual hashes are usually employed~\cite{li2010detecting}. Depending on the hashing function used, binary hashes are cheap to compute and store and robust to small perturbations such as color or resolution changes~\cite{venkatesan2000robust}. Furthermore, performing distance calculations for a nearest-neighbors search is more computationally efficient when using small hashes~\cite{lin2015deep}.

\par An \textit{adversarial image attack} is an attack that ``perturbs" an image in such a way that some target classifier or detector fails when given the image as input~\cite{goodfellow2014explaining}. These attacks occur in either a white box setting, where an attacker has full access to a model~\cite{moosavi2016deepfool}, or a gray- or black-box setting, where the adversary has varying levels of access to the target model, either in terms of query access or some general knowledge of the target model's architecture~\cite{ilyas2018black}. The perturbations themselves are computed via stochastic gradient descent that treats the input image as a variable to optimize.

\par The desired failure mode can either be \textit{evasion}~\cite{biggio2013evasion}, where an attacker seeks to perturb an image by as little as possible so that a classifier classifies the image as anything except its true class. A more difficult form of attack is a \textit{targeted attack}, where the image is perturbed so that it is classified as the desired target class. A still stronger outcome is a \textit{targeted collision attack}~\cite{shafahi2018poison}, where an image is perturbed so that from the point of view of a model, the adversarial image is identical to another target image. 

\par In this work, we target image hashing methods in general. Image hashes are typically computed in one of two ways - either via a deep hashing algorithm~\cite{liu2016deep,zhao2015deep}, where the embeddings from the penultimate layer of a deep network are encoded and binarized, or with what we denote as a ``shallow" hashing function, which uses deterministic image operations to produce a fixed-length hash~\cite{kozat2004robust,mihccak2001new,monga2006perceptual}. Previous work has focused on targeted or evasion attacks for deep hashes~\cite{li2020towards,yang2018adversarial}. The targeted attacks in~\cite{bai2020targeted,li2020towards} explore poisoning attacks for deep hashing systems, but optimize for a general target class and not the stronger criterion of an exact collision. Furthermore, for large scale production systems shallow hashes are often used as they contain far fewer parameters than a deep network and can be quickly computed, even on a CPU. In addition, they are extremely precise and are particularly well-suited for duplicate image detection. For greater applicability, we experiment with \textit{both} shallow and deep hashes, and show that shallow hashes are just as vulnerable as deep hashes to gradient-based adversarial attacks.

\par Collision attacks have been well studied in the field of cryptography~\cite{liang2007improved,stevens2006fast,wang2005collision}. The specific attack family we explore are \textit{preimage} attacks~\cite{rogaway2004cryptographic}, where an adversary seeks to create a message that has a specific hash value. Ideally, a hash function with $k$ bits requires an $O(2^k)$ amount of work to craft a collision, which is infeasible for typical hash lengths on the order of hundreds of bits. Technically, in this work we focus on \textit{second preimage} attacks, where we seek to perturb an image $x$ with noise $r$ such that it has the same hash as a target image $y$, i.e. $h(x+r) = h(y)$. By design, image hashes are robust under small perturbations, and do not exhibit the \textit{avalanche effect}~\cite{feistel1973cryptography} of cryptographically-secure hashes, and thus they are particularly vulnerable to second preimage attacks. In addition, an attacker can exploit the fact that image hashing algorithms are (mostly) differentiable, so a gradient-based method such as~\cite{Szegedy2013} gives an efficient way to find a collision.

Finally, the specific attack scenario we simulate is a \textit{poisoning} attack~\cite{hao2020adversarial} where a bank of image hashes is unknowingly polluted with an image designed to disrupt the duplicate image detection service. For instance, a bank may contain images that are not allowed to be posted by a third party, and any uploaded image is checked against this bank and flagged if there is a match. An adversary can imperceptibly perturb an image that would normally be placed in this bank (either automatically or via human intervention), but has a hash that matches a benign target image. After poisoning, all live instances of the benign image will then be either incorrectly removed or flagged. We explore the poisoning susceptibility of shallow and deep hashing systems in under various levels of model access.

%% file: method.tex
\section{Attack methods}

\par To perform a hash collision, one can modify a source image $x_i$ so that it has exactly the same hash as a target image $y_i$, according to some hash function ${h(x) = f^n \circ \ldots \circ f^1(x)}$, where each individual $f^i$ is a distinct step in the hashing process. Given the shorthand definition $f^i(x) = f^i \circ \ldots \circ f^1(x)$, the set of image pairs that will have identical hashes is defined as:
\begin{equation}
\mathcal{C} = \left\{ (x, y) : f^i(x) = f^i(y), i \leq n, x \neq y \right\}
\label{eq:collision_set}
\end{equation}

Thus given a target image $y$, our task is to find an attack image $x$ such that $(x, y) \in \mathcal{C}$. As it is intractable to search this set in a brute-force manner, we instead can instead approximate $x$ via gradient descent. We adopt a modified form of the box-constrained optimization method of \cite{Szegedy2013}, and for a given source and target image, we seek to derive a minimally perturbed image $x + r$ that minimizes a hash distance by optimizing Eq.~\ref{eq:attack_objective}.

\begin{equation}
\begin{aligned}
& \underset{r}{\text{minimize}}
& &  c \cdot ||r||_2^2\\
& \text{subject to}
& & |h(x + r) - h(y)| \leq d\\
&
& & x + r \in [0, 1]^m
\end{aligned}
\label{eq:attack_objective}
\end{equation}

Although Eq.~\ref{eq:attack_objective} admits optimizing for a general distance threshold $d$, for most experiments we set $d=0$ to encourage exact collisions. The hyperparameter $c$ controls the rate of change between the magnitude of the noise and the hash difference. The $L_1$ norm is used as a hash distance metric as we experiment exclusively with binary hashes.

\subsection{Interior surjective attacks}
A key observation regarding image hashing systems is that, if at any point in the hashing pipeline there exists a collision between two intermediate outputs produced by two different images, then all successive steps will also produce collisions. Eq.~\ref{eq:collision_set} does not restrict attacks to only colliding the final output, but also permits optimizing earlier steps in the hashing pipeline. For instance, in a deep hash, colliding one set of intermediate feature maps will result in a collision at the output of the deep network (as all operations after that point are identical). Attacking earlier stages in a hashing function may lead to better transfer attacks, a hypothesis we examine in the Experiments section. A variant of this method was used in~\cite{lu2020enhancing} (although not within the hashing domain), where the interior feature maps of a deep network were used as attack objectives.

Secondly, hashing functions by design are \textit{surjective} (also known as \textit{onto}). An image of arbitrary size consisting of $m$ pixel values is mapped to a binary hash of fixed length $k$ by the hashing function $h: \mathbb{R}^m \rightarrow \{0, 1\}^k$. By the pigeonhole principle, many inputs are mapped to the same hash, and it may be the case that semantically dissimilar images have identical hashes. Additionally, because a hash is small relative to the size of the input space, the magnitude of the perturbation $r$ required for a hash collision may be smaller than that needed for a different adversarial task. Because the overall hashing function is surjective, then there must exist an interior function $f^i$ that is also surjective, and sweeping over the interior functions of a hashing algorithm may uncover a particularly susceptible weak point. Finally, there are typically more informative gradients before the final output of the hashing algorithm, so optimization is easier if an interior function is targeted. Thus, a generalization of the optimization method of \cite{Szegedy2013} is given in Eq.~\ref{eq:surj_attack_objective}, where an additional minimization term is added to minimize the distance of one or a set of interior functions $\mathcal{I}$. 

\begin{equation}
\begin{aligned}
& \underset{r}{\text{minimize}}
& & \sum_{i\in\mathcal{I}}||f^i(x + r) - f^i(y)||_2^2 + c \cdot ||r||_2^2 \\
& \text{subject to}
& & |h(x + r) - h(y)| \leq d\\
&
& & x + r \in [0, 1]^m
\end{aligned}
\label{eq:surj_attack_objective}
\end{equation}

\begin{figure}[ht!]
    \centering
    \includegraphics[width=0.95\columnwidth]{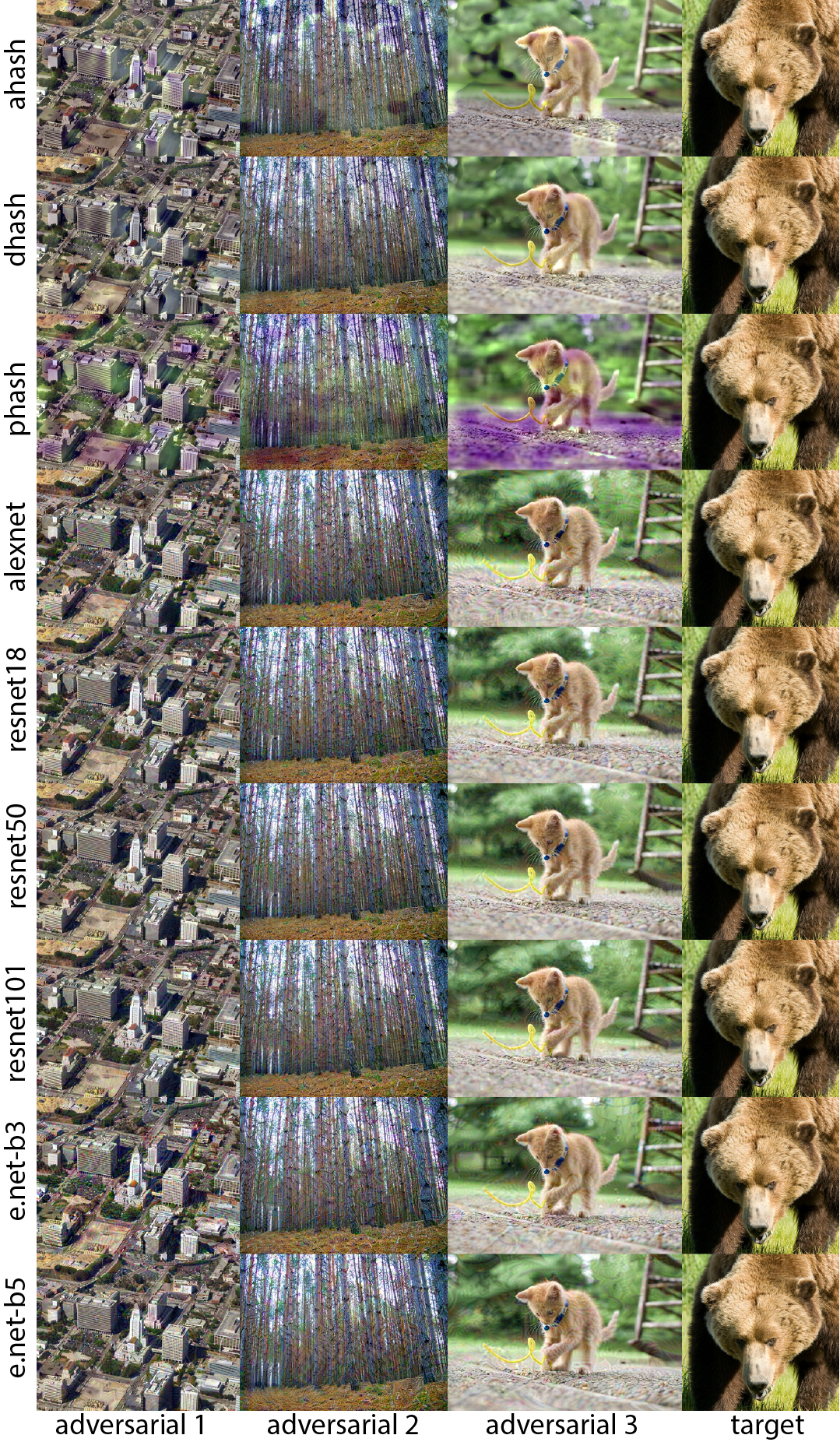}
    \caption{Results when optimizing the hinge loss in Eq.~\ref{eq:hash_hinge}}
    \label{fig:wb_hinge_results}
\end{figure}

Since binarization is common to most image hashing techniques, we can design a special-case surjective attack that enables small perturbations to have a large effect on the output hash. The last function $f^n$ in a hashing algorithm is usually implemented as a step function over each bit $j$:
\begin{equation}
h_j(x) = \begin{cases}
1, & \left(f^{n-1} \circ \ldots \circ f^0(x)\right)_j \geq 0\\
0, & \text{otherwise} 
\end{cases}
\end{equation}
The gradient of the step function $f^n$ is zero everywhere except zero, therefore optimization is difficult. However, $h_j(x)$ must only be perturbed by the minimum amount needed to push its output from either than zero to greater than or equal to zero (or vice-versa), so we can utilize a hinge loss to reflect this fact. As shown in Eq.~\ref{eq:hash_hinge}, we can also replace the step function itself with a sigmoid that acts like a soft step (similar to~\cite{yang2018adversarial}, which uses a $\tanh$ function) and contains useful gradients, especially around zero. 
\newpage
\begin{equation}
\sum_{j=0}^k \text{max}\left(0, |h_j(y) - \sigma(f^{n-1}_j(x + r))| - \delta\right) 
\label{eq:hash_hinge}
\end{equation}
In Eq~\ref{eq:hash_hinge}, $\delta < 0.5$ is a hyperparameter that controls the loss margin - smaller values of $\delta$ will push bits farther from the binarization boundary (potentially increasing transferability of the attack), while larger values will minimize the required perturbation of the original image. In our experiments, $\delta$ was fixed to 0.45 (w.l.o.g.).

\subsection{Hashing algorithms analyzed}\label{sec:hashes}
We examine not only the typical deep hashing methods already covered in some prior work, but a set of ``shallow" hashes. We define shallow hashes as those that do not use deep networks and do not contain learned components. Rather, they are entirely deterministic algorithms that typically consist of resampling an image to a fixed size and then running some other form of decimation to reduce the image further into a binary hash. 

Furthermore, we include image pre-processing steps as part of the overall hashing function and include them in the gradient-based adversarial attack optimization. Most adversarial attacks in the literature operate on fixed-size images, so the resizing algorithm is ignored in the optimization. However, image hashes are sensitive to the resizing algorithm used, and some previous work has examined the behavior of general adversarial attacks under different forms of resizing or compression~\cite{guo2017countering,prakash2018protecting,quiring2020adversarial,shin2017jpeg}. Accordingly, we directly target the resize step as a potential attack point. Furthermore, this allows for general attacks using any image as an adversarial or target image.

The specific shallow hashes we examined were variants of the methods described in~\cite{krawetz2020}, specifically aHash, dHash, and pHash. The methods are available as an open source Python package ImageHash\footnote{https://pypi.org/project/ImageHash/}. All three methods first converts the image to a single luma channel and then uses Lanczos resampling to resize an image to a fixed size. aHash computes the average over the resized image (either 8x8, 12x12, or 16x16, depending on the hash size) and uses this as a threshold to binarize each pixel into 64, 144, or 256 bits. dHash is similar, but resizes to a fixed size with an extra column of pixels, computes the horizontal image gradient of the resized image, and sets bits based on whether each gradient value is positive or negative. Finally, pHash performs a DCT on the resized image (which is set to twice the size of an aHash image of the same size), discards the high frequency components, and produces a bit based on whether or not the low-frequency DCT component exceeds or is below the median component value. For each hash, the resize function is similar to the other hashes, and provides a potential weak point to attack.
\par For deep hashes, we examined a variety of architectures developed over the past decade. These architectures were AlexNet~\cite{krizhevsky2012imagenet}, ResNet~\cite{he2016deep}, and EfficientNet~\cite{tan2019efficientnet}. AlexNet uses standard convolutional operations in a feedforward fashion; ResNet adds skip connections that propagate information forward to avoid vanishing gradients; EfficientNet currently provides the best performance on the ImageNet benchmark\footnote{https://sotabench.com/benchmarks/image-classification-on-imagenet} and uses fewer parameters than an equivalent ResNet, which makes it well suited for mobile or production environments. Each was trained on the ImageNet dataset. For each deep network, the hash computation step was performed with PCA followed by binarization. PCA was performed on all embeddings in the ImageNet training set, and a set of components designed to project embeddings to 64, 128, and 256 dimensions were computed. Then, the median of each vector element of the projected embeddings was computed and saved. Finally when computing the binary hash at inference time, the inferred embedding is projected and binarized according to whether or not a value at a given element was greater than or equal to the stored median value.

%% file: experiments.tex
\section{Experiments}
We performed a variety of white box, gray box, and transfer attacks attacks against the hashes listed in Sec.~\ref{sec:hashes}. In the white box setting, we consider only an exact hash collision as an attack ``success." Although duplicate image detection systems usually use a small distance threshold for finding ``exact" matches, exact collisions allow an attacker to avoid having to determine the match threshold through repeated attacks. The white box setting is designed to simulate how successful an attacker might be in poisoning a detection system if they knew the algorithm used to produce the hash bank used in that system (for instance, if the system used an open source hash). For gray and black box settings, because these hash systems often use a detection threshold, we provide additional analysis on the attack success rates when using a precision-tuned distance threshold greater than zero. Only 256 bit hash results are shown here, but results for 64 and 128 bit hashes are given in the supplementary material.

For each hash type and architecture, we pre-defined a set of ``split points" where a particular interior function was used as an optimization objective as shown in Eq.~\ref{eq:surj_attack_objective}. For the shallow hashes, these split points were after every step in the hashing process, and so the total number varied depending on the algorithm used. For the deep hashes, 9 split points were chosen at roughly equal depths across each architecture, and each step in the PCA hashing function was also chosen as a split point, giving an overall total of 12 split points per deep network. For AlexNet, the split points were placed after ReLU layers; for the ResNet, after residual blocks; for EfficientNet, after equally-spaced mobile invertable convolutional blocks.

A final consideration is the choice of source images when attacking a given target image. It may be the case that it is easier to perturb semantically-related images, especially in the case where deep hases are used, as deep networks are usually trained for a particular classification task and then co-opted for use in image hashing. To this end, for each model, split point, and hash length combination, we tested 1,000 random image pairs from different ImageNet classes and 1,000 image pairs taken from the same ImageNet class in order to determine if there is an advantage in using similar source and target images. Also note that all visualized images in this paper were \textit{not} sampled from ImageNet, which provides qualitative evidence that this attack framework is agnostic to specific training distributions.

\subsection{Baselines}\label{sec:exp_baselines}
Prior to measuring our adversarial hash attack success rates, we computed a set of baselines over each hash to give a rough idea of the performance of each under light noise and perturbations. To compute this, we used a perturbed version of the ImageNet validation set as probes, and performed a nearest-neighbors lookup into the unperturbed validation set using FAISS~\cite{johnson2019billion}. Furthermore, we used the original training images as a set of distractors. The top-k accuracy of each hash is shown in Table~\ref{tab:baselines}. Additionally, to verify that our attacks are not measuring spurious successes, we computed a baseline collision rate which measures how likely it is for a random image to share the exact hash as another image. For brevity, only 256-bit hashes are included in Table~\ref{tab:baselines}, but full results are available in the supplemental material. The maximum incidental collision rate was 0.039 for the 64-bit version of aHash; most hashes besides the simplistic aHash had a baseline collision rate less than 0.001.

\begin{table}
\setlength{\tabcolsep}{3pt}
\centering
\scriptsize
\begin{tabular}{|l|rrrrrr|}
\hline
 Hash  & Top-1-acc &   Top-10-acc & Coll. rate & Succ. rate & $L_2$ loss & SSIM\\
\hline
 ahash\_256 & 0.611 & 0.628 & 0.002 & 1.000 & 0.032 & 0.759       \\
 dhash\_256 & 0.617 & 0.632 & 0.000 & 0.970 & 0.022 &       \\
 phash\_256 & 0.614 & 0.624 & 0.000 & 0.973 & 0.030 & -       \\
 a.net\_256 & 0.770  & 0.847 & 0.000 & 0.941 & 0.004 & 0.608       \\
 r.net18\_256 & 0.868 & 0.926 & 0.000 & 1.000 & 0.001 & 0.664            \\
 r.net50\_256 & 0.864 & 0.933 & 0.000 & 1.000 & 0.001 & 0.657           \\
 r.net101\_256 & 0.869 & 0.936 & 0.000 & 1.000 & 0.001 & 0.661           \\
 e.net-b3\_256 & 0.876 & 0.934 & 0.000 & 0.957 & 0.001 & 0.623        \\
 e.net-b5\_256 & 0.838 & 0.908 & 0.000 & 1.000 & 0.001 & 0.630        \\
\hline
\end{tabular}
\vspace*{2mm}
\caption{Baseline nearest-neighbor classification accuracy for all 256-bit hashes, as well as incidental collision rates, exact collision attack success rates with the hinge loss, and the average $L_2$ content loss and SSIM~\cite{wang2004image} at the exact collision.}
\label{tab:baselines}
\end{table}


\begin{figure*}[ht!]
    \centering
    \includegraphics[width=\textwidth]{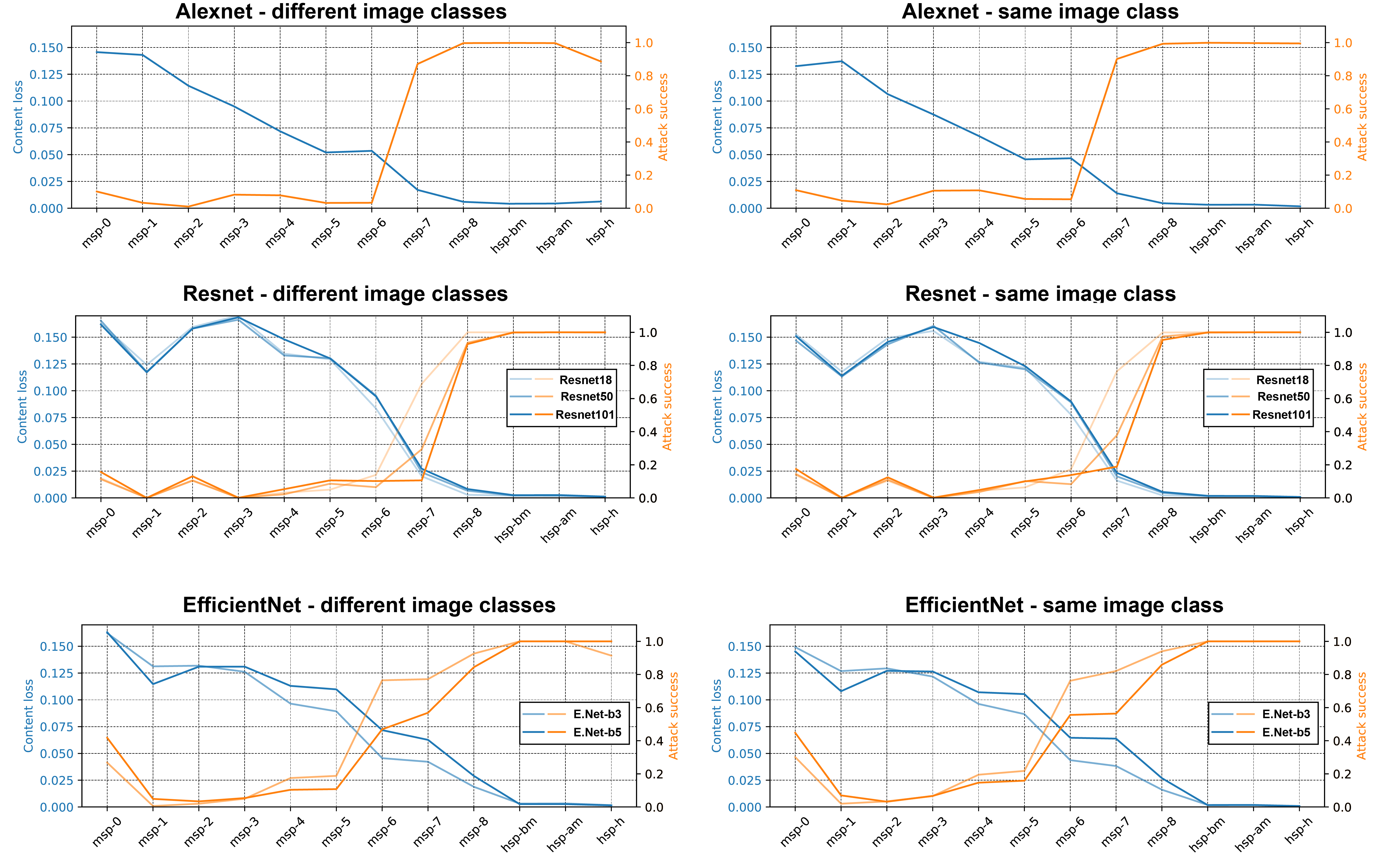}
    \caption{White box attack success rates at different split points for various deep network architectures. Earlier split points require larger adversarial perturbations and give better gray box success rates. Later points are effective when targeting a known model.}
    \label{fig:deep_wb_success_graph}
\end{figure*}

\subsection{White box attacks}
In the white box setting, an attacker has full access to a model or algorithm and any parameters to that algorithm. To perform this set of white box attacks, we divided the ImageNet validation set into two partitions of 500 classes each, the first of which was used for hyperparameter tuning, and the second for measuring attack success. The validation set was used as it contains class labels that can be used to measure if semantically-similar images are easier to collide. For each of the 1,000 random pairs and 1,000 class pairs, we saved the output white box adversarial images for testing our approach's gray and black box performance. For each model and hash type, we optimized Eq.~\ref{eq:surj_attack_objective} using  the Adam optimization method (with betas 0.1, 0.1). The learning rates were determined via hyper-parameter tuning (5.0 for shallow hashes, and 0.005 for deep hashes), and the content loss weight $c$ was fixed at 0.001 for all experiments in order to maximize the chance of achieving a hash collision.

\begin{figure}
    \centering
    \includegraphics[width=\columnwidth]{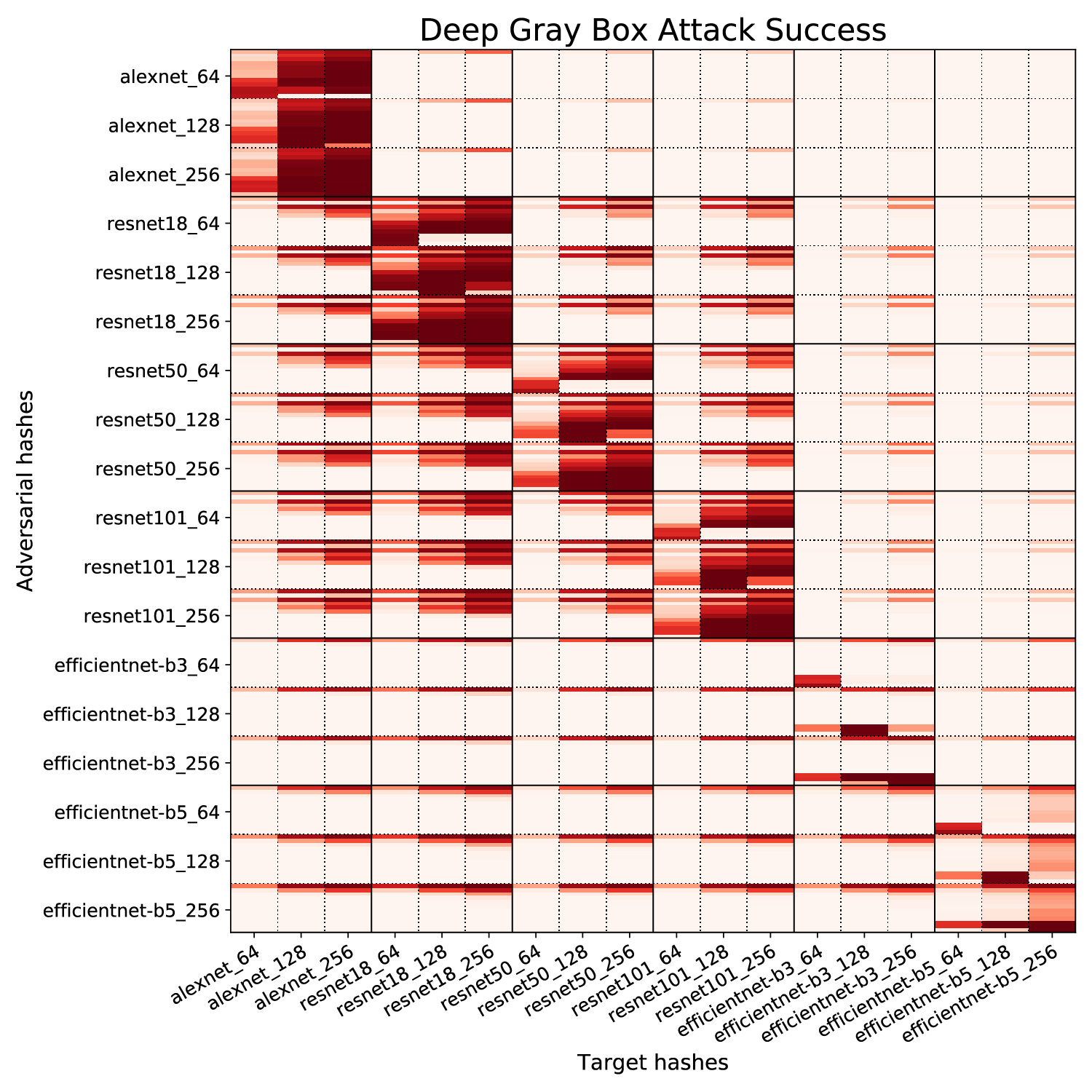}
    \caption{Deep gray box attacks - darker color corresponds to better success. Vertical axis shows the hash targeted for optimization, with higher bins in each box corresponding to earlier split points. The horizontal axis shows the target hashes for the gray box attack.}
    \label{fig:deep_gb}
\end{figure}

\begin{figure}
    \centering
    \includegraphics[width=\columnwidth]{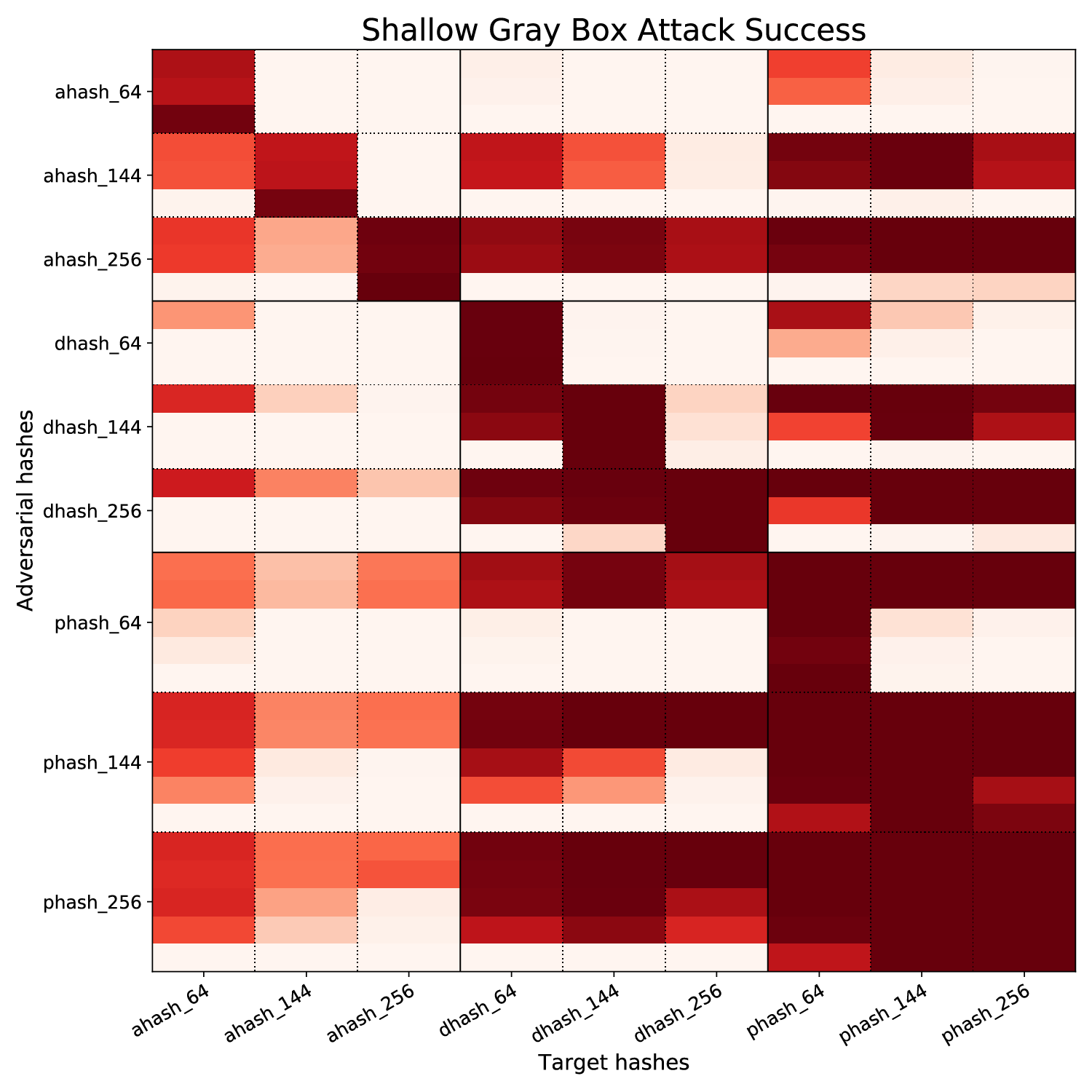}
    \caption{Shallow gray box attacks, with same structure as Fig.~\ref{fig:deep_gb}.\vspace*{-4mm}}
    \label{fig:shallow_gb}
\end{figure}

Fig.~\ref{fig:wb_hinge_results} demonstrates the white box capabilities of optimizing with the hinge loss in Eq.~\ref{eq:hash_hinge}. Each row shows a set of adversarial images, each of which has the exact same hash as the target image, generated by optimizing over a particular deep or shallow hash. Clearly if the parameters and algorithm of a hashing system are known, then white box attacks are extremely effective.

\begin{figure*}[h!]
    \centering
    \includegraphics[width=0.96\textwidth]{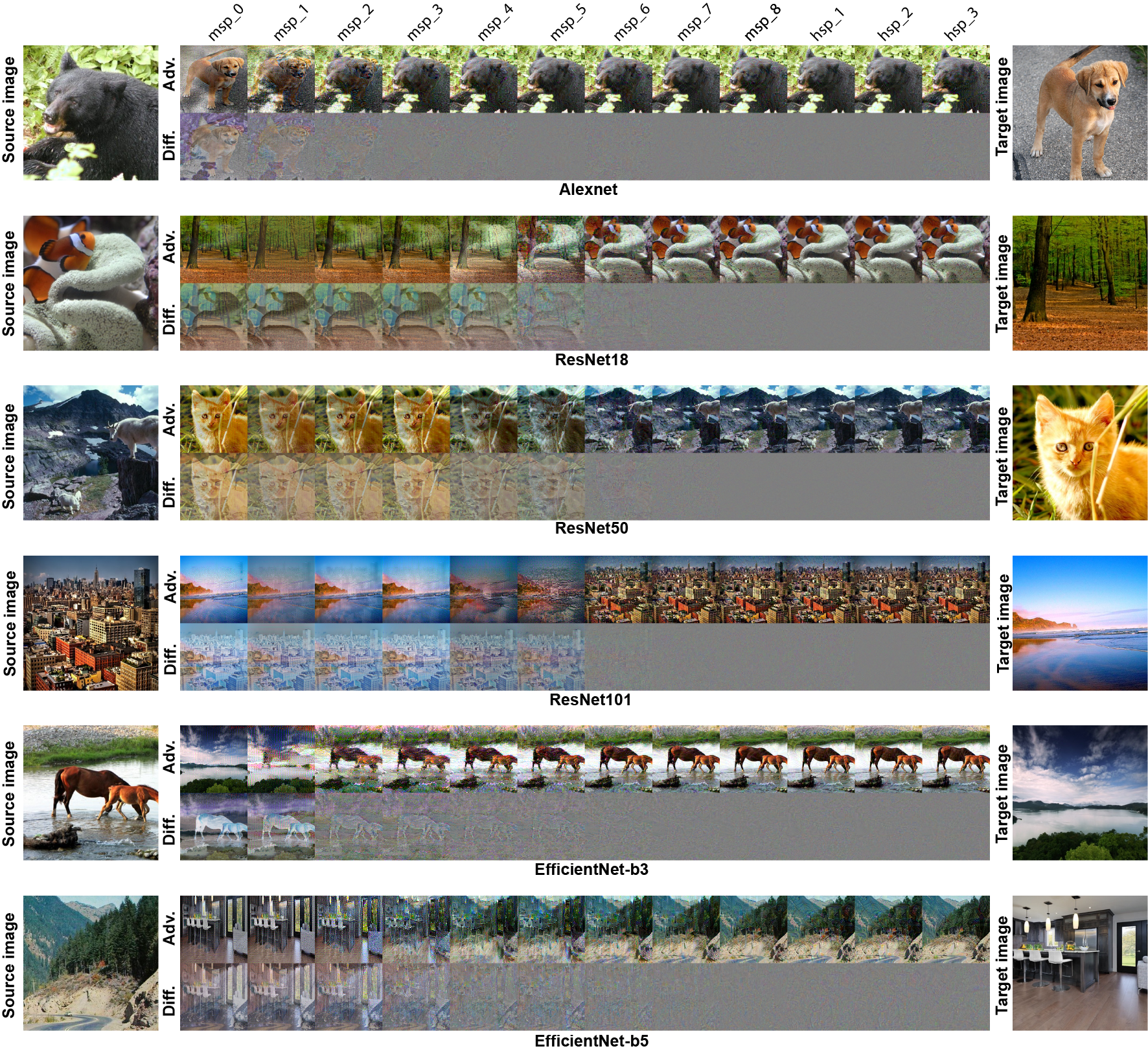}
    \caption{Visualized images for deep hashes over different split points.}
    \label{fig:deep_wb_depth_ims_1}
    \vspace*{-1mm}
\end{figure*}

Fig.~\ref{fig:deep_wb_success_graph} demonstrates that when using a source image that is semantically related to the target image, the attacks are slightly more successful and with less perturbation required. In addition, the figure shows that for deep networks, there is a clear inflection point as we attack later points in the network. In fact as shown in Fig.~\ref{fig:deep_wb_depth_ims_1}, images produced by attacking earlier layers in deep networks tend to be mostly semantically transformed into the target image itself (thus defeating the purpose of the attack). However, especially around model split points 7 and beyond, the attack is much more effective at preserving the original source image while still producing an exact hash collision.  The most likely explanation is that it is at this point in the network where lower-level features are combined into hierarchical semantic concepts. Specifically for the ResNet architecture, the skip connections propagate the original image deep into the network, providing a good defense against this attack. A corollary of this result showing inflection points is that this may be the optimal point to fine tune a vision model for transfer learning.

\begin{figure}
    \centering
    \includegraphics[width=\columnwidth]{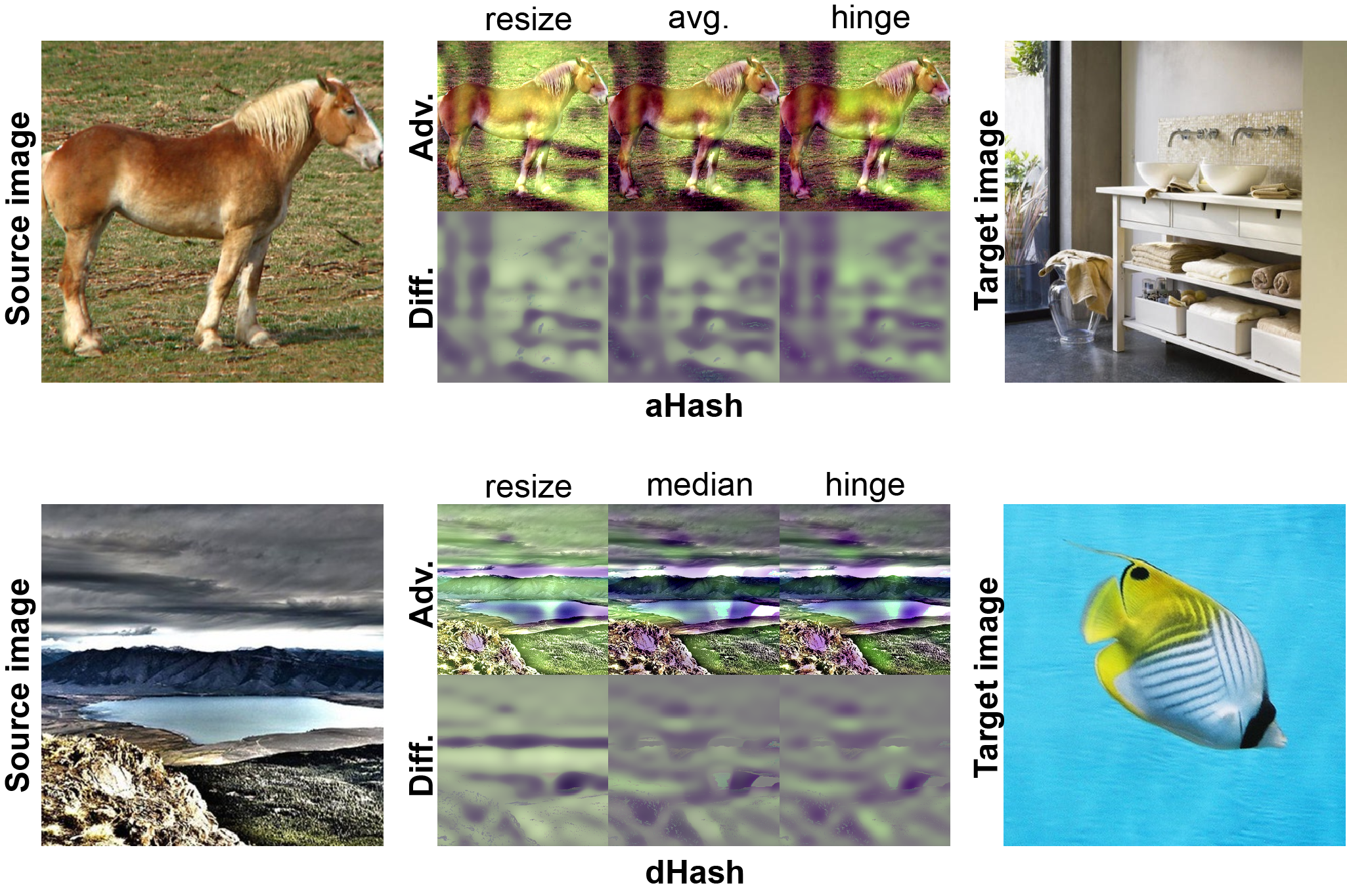}
    \caption{Visualized images for shallow hashes over different split points.\vspace*{-5mm}}
    \label{fig:shallow_depth}
\end{figure}

\subsection{Gray box attacks}
In the gray box setting, an attacker knows some information about the hashing algorithm used. For instance, the adversary may know that a pre-trained ResNet is used to extract primary feature embeddings, but they do not know the algorithm used to produce the final binary hash. To measure the efficacy of our approach, we grouped the shallow hashes together, as well as the deep models from the same architectural family, and measured every gray box pair. 

As seen in Fig.~\ref{fig:deep_wb_success_graph}, for some early split points it is difficult to achieve exact hash collisions in a white box setting, so instead we measure the success rate at a pre-computed "exact match" distance threshold. These thresholds were derived for each hash by using the augmented images described in Sec.~\ref{sec:exp_baselines}, and finding the distance such that the nearest neighbor in the given hash space was an exact match with precision 0.99 (these threshold values are given in the supplementary material). Then for any adversarial image/target image pair, if their distance is less than this threshold, the attack is marked as a success. Within each cell in Figs.~\ref{fig:deep_gb} and \ref{fig:shallow_gb}, the success rate is plotted, with higher values in each cell corresponding to an earlier attack point for the adversarial hash. As is evident in Fig.~\ref{fig:deep_gb}, our attack transfers well between different hash functions that use similar architectures, and transfers especially well if the same base architecture is used, regardless of the hashing function appended to that architecture. However, transferability decreases as deeper split points are used.

For shallow hashes, all three share similar resizing functions, so it is a particularly suitable interior surjective attack point for gray box attacks. Fig.~\ref{fig:shallow_gb} demonstrates that if an adversary were to \textit{only} know the algorithm used to perform a resize, and were able to produce a collided resized image, the downstream hashing method \textit{does not matter}.

\subsection{Transfer attacks}
Finally, we measured the ability of an adversary to use an unrelated hash to create collisions - specifically if a different deep architecture was used, or a collision was optimized for a shallow hash and then applied to a deep hash (and vice-versa). In the supplementary material, a combined version of Figs.~\ref{fig:deep_gb},~\ref{fig:shallow_gb} are provided. The transfer attack success rate was very low, indicating that shallow and deep hashes differ significantly and that a shallow/deep hash is ``safe" from an attack targeting a deep/shallow hash. Note that while ImageNet was solely used as it is fairly representative of the distribution of ``all images," different classification tasks may produce different low-level convolutional filters. Additionally we observed diminishing returns across different deep architecture types that were trained on the same dataset, so we surmise that using networks trained on a secret dataset or keeping the architecture secret provides an appropriate defense to hash collision attacks.

%% file: mitigations.tex
\section{Mitigations and conclusions}
Based on our comprehensive analysis, several conclusions can be drawn. First, all attacks that have white box access to a model are able to produce exact hash collisions with minimal perturbations of the source image. Consequently, it is necessary that the algorithm used in a production system is not made public if one wishes to guarantee the security of said system - otherwise the system is completely vulnerable to poisoning attacks. Second, we demonstrate that our attack method is successful even in a gray box setting, so one must withhold even general details about the algorithm such as the training set or base model. A third approach is more practical than the above methods which rely on ``security by obscurity." Transfer attacks between shallow and deep hashes were much less successful, indicating that the image operations in both hashing algorithms are fundamentally different. Thus using a combination shallow/deep hash will provide more security as an adversarial attack must perturb an image to fool both models that operate in an essentially orthogonal manner.

%% file: main.bbl
\begin{thebibliography}{10}\itemsep=-1pt

\bibitem{photodna}
{PhotoDNA}.
\newblock https://www.microsoft.com/en-us/photodna.

\bibitem{babenko2016efficient}
A.~Babenko and V.~Lempitsky.
\newblock Efficient indexing of billion-scale datasets of deep descriptors.
\newblock In {\em Conference on Computer Vision and Pattern Recognition
  (CVPR)}, pages 2055--2063, 2016.

\bibitem{bai2020targeted}
J.~Bai, B.~Chen, Y.~Li, D.~Wu, W.~Guo, S.-t. Xia, and E.-h. Yang.
\newblock Targeted attack for deep hashing based retrieval.
\newblock {\em arXiv preprint arXiv:2004.07955}, 2020.

\bibitem{biggio2013evasion}
B.~Biaggio, I.~Corona, D.~Maiorca, B.~Nelson, N.~{\v{S}}rndi{\'c}, P.~Laskov,
  G.~Giacinto, and F.~Roli.
\newblock Evasion attacks against machine learning at test time.
\newblock In {\em Joint European Conference on Machine Learning and Knowledge
  Discovery in Databases}, pages 387--402. Springer, 2013.

\bibitem{chum2008near}
O.~Chum, J.~Philbin, A.~Zisserman, et~al.
\newblock Near duplicate image detection: min-hash and tf-idf weighting.
\newblock In {\em BMVC}, volume 810, pages 812--815, 2008.

\bibitem{dong2012high}
W.~Dong, Z.~Wang, M.~Charikar, and K.~Li.
\newblock High-confidence near-duplicate image detection.
\newblock In {\em ACM International Conference on Multimedia Retrieval}, pages
  1--8, 2012.

\bibitem{feistel1973cryptography}
H.~Feistel.
\newblock Cryptography and computer privacy.
\newblock {\em Scientific American}, 228(5):15--23, 1973.

\bibitem{goodfellow2014explaining}
I.~J. Goodfellow, J.~Shlens, and C.~Szegedy.
\newblock Explaining and harnessing adversarial examples.
\newblock {\em arXiv preprint arXiv:1412.6572}, 2014.

\bibitem{guo2017countering}
C.~Guo, M.~Rana, M.~Cisse, and L.~Van Der~Maaten.
\newblock Countering adversarial images using input transformations.
\newblock {\em arXiv preprint arXiv:1711.00117}, 2017.

\bibitem{Integrity2020}
A.~Halevy, C.~{Canton Ferrer}, H.~Ma, U.~Ozertem, P.~Pantel, M.~Saeidi,
  F.~Silvestri, and V.~Stoyanov.
\newblock Preserving integrity in online social networks.
\newblock {\em arXiv preprint arXiv:2009.10311}, 2020.

\bibitem{hao2020adversarial}
H.~X. Y.~M. Hao-Chen, L.~D. Deb, H.~L. J.-L.~T. Anil, and K.~Jain.
\newblock Adversarial attacks and defenses in images, graphs and text: A
  review.
\newblock {\em International Journal of Automation and Computing},
  17(2):151--178, 2020.

\bibitem{he2016deep}
K.~He, X.~Zhang, S.~Ren, and J.~Sun.
\newblock Deep residual learning for image recognition.
\newblock In {\em Conference on Computer Vision and Pattern Recognition
  (CVPR)}, pages 770--778, 2016.

\bibitem{ilyas2018black}
A.~Ilyas, L.~Engstrom, A.~Athalye, and J.~Lin.
\newblock Black-box adversarial attacks with limited queries and information.
\newblock {\em arXiv preprint arXiv:1804.08598}, 2018.

\bibitem{johnson2019billion}
J.~Johnson, M.~Douze, and H.~J{\'e}gou.
\newblock Billion-scale similarity search with gpus.
\newblock {\em IEEE Transactions on Big Data}, 2019.

\bibitem{ke2004efficient}
Y.~Ke, R.~Sukthankar, L.~Huston, Y.~Ke, and R.~Sukthankar.
\newblock Efficient near-duplicate detection and sub-image retrieval.
\newblock In {\em ACM Multimedia}, volume~4, page~5. Citeseer, 2004.

\bibitem{kozat2004robust}
S.~S. Kozat, R.~Venkatesan, and M.~K. Mih{\c{c}}ak.
\newblock Robust perceptual image hashing via matrix invariants.
\newblock In {\em International Conference on Image Processing (ICIP)},
  volume~5, pages 3443--3446, 2004.

\bibitem{krawetz2020}
N.~Krawetz.
\newblock Looks like it.
\newblock
  \url{http://hackerfactor.com/blog/index.php\%3F/archives/432-Looks-Like-It.html}.
\newblock Accessed: 2020-10-26.

\bibitem{krizhevsky2012imagenet}
A.~Krizhevsky, I.~Sutskever, and G.~E. Hinton.
\newblock {ImageNet} classification with deep convolutional neural networks.
\newblock In {\em Advances in Neural Information Processing Systems (NIPS)},
  pages 1097--1105, 2012.

\bibitem{li2020towards}
M.~Li, C.~Deng, T.~Li, J.~Yan, X.~Gao, and H.~Huang.
\newblock Towards transferable targeted attack.
\newblock In {\em Conference on Computer Vision and Pattern Recognition
  (CVPR)}, pages 641--649, 2020.

\bibitem{li2010detecting}
M.~Li, B.~Wang, W.-Y. Ma, and Z.~Li.
\newblock Detecting duplicate images using hash code grouping, Jan.~12 2010.
\newblock US Patent 7,647,331.

\bibitem{liang2007improved}
J.~Liang and X.-J. Lai.
\newblock Improved collision attack on hash function {MD5}.
\newblock {\em Journal of Computer Science and Technology}, 22(1):79--87, 2007.

\bibitem{lin2015deep}
K.~Lin, H.-F. Yang, J.-H. Hsiao, and C.-S. Chen.
\newblock Deep learning of binary hash codes for fast image retrieval.
\newblock In {\em Conference on Computer Vision and Pattern Recognition
  Worshops (CVPRW)}, pages 27--35, 2015.

\bibitem{liu2016deep}
H.~Liu, R.~Wang, S.~Shan, and X.~Chen.
\newblock Deep supervised hashing for fast image retrieval.
\newblock In {\em Conference on Computer Vision and Pattern Recognition
  (CVPR)}, pages 2064--2072, 2016.

\bibitem{lu2010forensic}
W.~Lu, A.~L. Varna, and M.~Wu.
\newblock Forensic hash for multimedia information.
\newblock In {\em Media Forensics and Security II}, volume 7541.

\bibitem{lu2020enhancing}
Y.~Lu, Y.~Jia, J.~Wang, B.~Li, W.~Chai, L.~Carin, and S.~Velipasalar.
\newblock Enhancing cross-task black-box transferability of adversarial
  examples with dispersion reduction.
\newblock In {\em Conference on Computer Vision and Pattern Recognition
  (CVPR)}, pages 940--949, 2020.

\bibitem{mihccak2001new}
M.~K. M{\i}h{\c{c}}ak and R.~Venkatesan.
\newblock New iterative geometric methods for robust perceptual image hashing.
\newblock In {\em ACM Workshop on Digital Rights Management}, pages 13--21.
  Springer, 2001.

\bibitem{monga2006perceptual}
V.~Monga and B.~L. Evans.
\newblock Perceptual image hashing via feature points: performance evaluation
  and tradeoffs.
\newblock {\em IEEE Transactions on Image Processing}, 15(11):3452--3465, 2006.

\bibitem{moosavi2017universal}
S.-M. Moosavi-Dezfooli, A.~Fawzi, O.~Fawzi, and P.~Frossard.
\newblock Universal adversarial perturbations.
\newblock In {\em Conference on Computer Vision and Pattern Recognition
  (CVPR)}, pages 1765--1773, 2017.

\bibitem{moosavi2016deepfool}
S.-M. Moosavi-Dezfooli, A.~Fawzi, and P.~Frossard.
\newblock Deepfool: a simple and accurate method to fool deep neural networks.
\newblock In {\em Conference on Computer Vision and Pattern Recognition
  (CVPR)}, pages 2574--2582, 2016.

\bibitem{prakash2018protecting}
A.~Prakash, N.~Moran, S.~Garber, A.~DiLillo, and J.~Storer.
\newblock Protecting jpeg images against adversarial attacks.
\newblock In {\em 2018 Data Compression Conference}, pages 137--146. IEEE,
  2018.

\bibitem{quiring2020adversarial}
E.~Quiring, D.~Klein, D.~Arp, M.~Johns, and K.~Rieck.
\newblock Adversarial preprocessing: Understanding and preventing image-scaling
  attacks in machine learning.
\newblock In {\em {USENIX} Security Symposium}, 2020.

\bibitem{rogaway2004cryptographic}
P.~Rogaway and T.~Shrimpton.
\newblock Cryptographic hash-function basics: Definitions, implications, and
  separations for preimage resistance, second-preimage resistance, and
  collision resistance.
\newblock In {\em International Workshop on Fast Software Encryption}, pages
  371--388. Springer, 2004.

\bibitem{shafahi2018poison}
A.~Shafahi, W.~R. Huang, M.~Najibi, O.~Suciu, C.~Studer, T.~Dumitras, and
  T.~Goldstein.
\newblock Poison frogs! targeted clean-label poisoning attacks on neural
  networks.
\newblock In {\em Advances in Neural Information Processing Systems (NIPS)},
  pages 6103--6113, 2018.

\bibitem{shin2017jpeg}
R.~Shin and D.~Song.
\newblock {JPEG}-resistant adversarial images.
\newblock In {\em Workshop on Machine Learning and Computer Security},
  volume~1, 2017.

\bibitem{steinebach2012forbild}
M.~Steinebach, H.~Liu, and Y.~Yannikos.
\newblock Forbild: Efficient robust image hashing.
\newblock In {\em Media Watermarking, Security, and Forensics 2012}, volume
  8303.

\bibitem{stevens2006fast}
M.~Stevens.
\newblock Fast collision attack on {MD5}.
\newblock {\em IACR Cryptol. ePrint Arch.}, 2006:104, 2006.

\bibitem{Szegedy2013}
C.~Szegedy, W.~Zaremba, I.~Sutskever, J.~Bruna, D.~Erhan, I.~Goodfellow, and
  R.~Fergus.
\newblock Intriguing properties of neural networks.
\newblock {\em arXiv preprint arXiv:1312.6199}, 2013.

\bibitem{tan2019efficientnet}
M.~Tan and Q.~V. Le.
\newblock {EfficientNet}: Rethinking model scaling for convolutional neural
  networks.
\newblock {\em arXiv preprint arXiv:1905.11946}, 2019.

\bibitem{venkatesan2000robust}
R.~Venkatesan, S.-M. Koon, M.~H. Jakubowski, and P.~Moulin.
\newblock Robust image hashing.
\newblock In {\em International Conference on Image Processing (ICIP)},
  volume~3, pages 664--666. IEEE, 2000.

\bibitem{venkatesan2003system}
R.~Venkatesan and S.-M.~W. Koon.
\newblock System and method for hashing digital images, Dec.~30 2003.
\newblock US Patent 6,671,407.

\bibitem{wang2005collision}
X.~Wang, Y.~Yin, and H.-G. Yu.
\newblock Collision search attacks on sha1.
\newblock {\em Crypto 2005}, 2005.

\bibitem{wang2004image}
Z.~Wang, A.~C. Bovik, H.~R. Sheikh, and E.~P. Simoncelli.
\newblock Image quality assessment: from error visibility to structural
  similarity.
\newblock {\em IEEE Transactions on Image Processing}, 13(4):600--612, 2004.

\bibitem{yang2018adversarial}
E.~Yang, T.~Liu, C.~Deng, and D.~Tao.
\newblock Adversarial examples for hamming space search.
\newblock {\em IEEE Transactions on Cybernetics}, 2018.

\bibitem{zhao2015deep}
F.~Zhao, Y.~Huang, L.~Wang, and T.~Tan.
\newblock Deep semantic ranking based hashing for multi-label image retrieval.
\newblock In {\em Conference on Computer Vision and Pattern Recognition
  (CVPR)}, pages 1556--1564, 2015.

\end{thebibliography}
